\documentclass[letterpaper,10pt,conference]{ieeeconf}

\IEEEoverridecommandlockouts 

\usepackage[T1]{fontenc}
\usepackage{amsmath}
\usepackage{booktabs}
\usepackage{graphicx}
\usepackage{tikz}
\usepackage[backend=biber,style=ieee,doi=false,url=false]{biblatex}
\usepackage[font=footnotesize]{caption}
\usepackage{hyperref}

\title{\bfseries
RoboFin3D: A Sim-to-Real Platform for Robotic Surface Finishing
}

\author{
  Haowei Wen$^{*}$,
  Shangtao Li$^{*}$,
  Vaibhav Sanjay,
  Philip Huang,
  Jiaoyang Li
  and Changliu Liu%
\thanks{$^{*}$Equal contribution.}%
\thanks{All authors are with the Robotics Institute,
School of Computer Science, Carnegie Mellon University.}%
}

\begin{document}

\maketitle

\begin{abstract}
Grinding and sanding are fundamental processes in industrial robotic surface finishing. 
However, physical trials are expensive and consume workpieces, making reproducible experiments difficult.
We present RoboFin3D, a sim-to-real platform built on Isaac Sim and the Newton physics engine,
that provides physics-based grinding and sanding simulation for cheap and repeatable robotic surface finishing experiments.
RoboFin3D utilizes a \emph{signed distance field} (SDF) to model the changing geometry of the workpiece, enabling contact computation, live updates and rendering without an intermediate mesh.
It additionally uses a separate \emph{surface field} to model progressive surface appearance change during sanding.
We also introduce WeldGen, a weld sampling module, to generate weld beads on 8,918 real-world workpiece meshes
for providing diverse simulation assets. 
The simulation parameters are calibrated on real experimental results
and our evaluation demonstrates our simulation's fidelity against the real world.
We also demonstrate that simulation-generated data can be used to improve the performance of perception models.
Simulation-only fine-tuning of SAM2 improves IoU for segmentation of unsanded regions from 77.15\% to 84.47\%,
while combined synthetic and real training reaches 97.41\%.
\end{abstract}

\begin{figure}[!t]
    \centering
    \resizebox{\columnwidth}{!}{\input{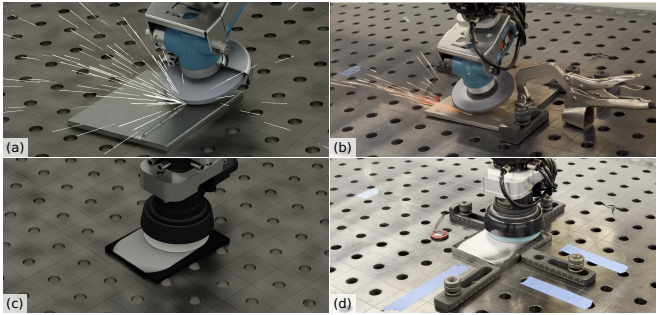}}
    \caption{
      RoboFin3D is able to model grinding material removal and sanding appearance changes.
      (a) and (b) are simulated and real grinding processes.
      (c) and (d) are simulated and real sanding processes.
    }
    \label{fig:teaser}
\end{figure}

\section{Introduction}

\begin{figure*}[!t]
  \centering
  \includegraphics[width=\textwidth]{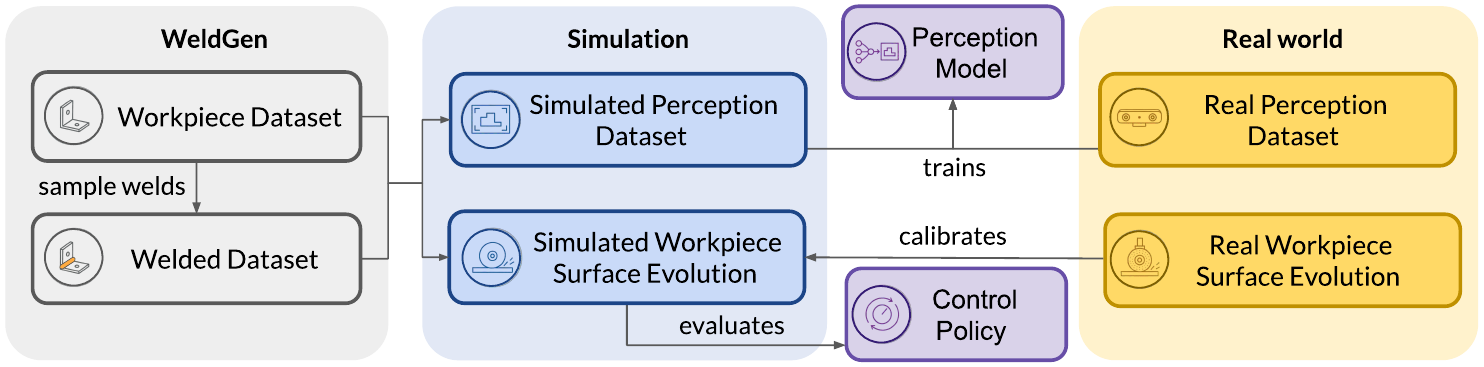}
  \caption{RoboFin3D workflow. WeldGen generates welds on workpieces.
  They are then used as both part of the training data for the perception model and assets in the simulator.
  Our perception model is fine-tuned using synthetic and real images.
  Real experimental results are used to calibrate the simulation,
  and the simulator can be used to evaluate a grinding or sanding control policy before real deployment.}
  \label{fig:platform}
\end{figure*}

Robotic surface finishing removes material from a workpiece with an abrasive
tool, so every physical trial changes the workpiece permanently. Developing
and evaluating a control policy takes many trials, but a processed workpiece cannot
be restored to its original starting state. Real-world experiments are therefore
expensive~\cite{hachimine2025csd} and hard to reproduce. A simulation platform
can restore a workpiece to any earlier state and rerun a trial at no material
cost, making the evaluation of robotic surface finishing reproducible,
cheaper, and easier.

Work related to robotic finishing spans physical systems, process simulation,
surface rendering, and perception. Robotic finishing systems integrate surface
reconstruction, motion planning, and controlled
execution~\cite{huo2019autonomous,alt2024robogrind}. Process models predict
belt-grinding outcomes~\cite{ren2006simulation}, and Cutting Sequence Diffuser
uses geometric material-removal simulation to learn grinding sequences that
transfer to real robots~\cite{hachimine2025csd}. For surface visualization,
Beer et al.\ render machining results using triangulated workpieces with
surface microtopography~\cite{beer2023rendering}. Visual monitoring assesses
weld-removal stages~\cite{pandiyan2019verification}, and physically
synthesized surface images support scratch segmentation with limited real
training data~\cite{wang2026scratch}. Evaluating finishing robots in
simulation requires robot contact, the resulting workpiece change, and
rendered camera images in one simulation, which none of these works provides.

We present RoboFin3D, a sim-to-real platform for robotic surface finishing.
RoboFin3D models two finishing processes (Fig.~\ref{fig:teaser}). Grinding
uses a rigid grinding wheel to remove large amounts of material, such as weld
beads. Sanding uses coated abrasives, such as sandpaper, mainly to smooth
surfaces or remove paint and rust. RoboFin3D simulates both processes in Isaac
Sim with the Newton physics engine. A signed distance field (SDF) records the
changing geometry of the workpiece, and a separate surface field records the
changing appearance of its surface. This representation supports hydroelastic
contact computation, live updates as the tool moves, and rendering without
intermediate mesh extraction. The physical parameters of the simulation are
calibrated from real grinding and sanding experiments.
Fig.~\ref{fig:platform} shows the RoboFin3D workflow.

Our contributions are:

\begin{enumerate}
  \item \textbf{Grinding and sanding simulation.}
  We develop a simulator in which contact computation, material removal,
  and rendering operate directly on the SDF and the surface field, so that
  changes to the workpiece affect subsequent contact and the rendered
  camera images.

  \item \textbf{Welded workpiece dataset.}
  We build a dataset of 8,918 welded workpieces from selected CAD models~\cite{koch2019abc}
  for evaluating surface finishing policies in simulation.
  WeldGen samples weld curves of controllable difficulty on each workpiece,
  adds bead geometry, and records the ground-truth weld geometry.

  \item \textbf{Evaluation of simulation fidelity and perception transfer.}
  We evaluate simulated grinding and sanding against physical trials
  and show that the simulated processing effects resemble real ones.
  We also demonstrate that fine-tuning a perception model on simulated images
  is able to improve its segmentation performance on real workpieces.
\end{enumerate}

\section{Simulation Platform}
\label{sec:simulation_platform}

\begin{figure}[!t]
    \centering
    \includegraphics[width=\columnwidth]{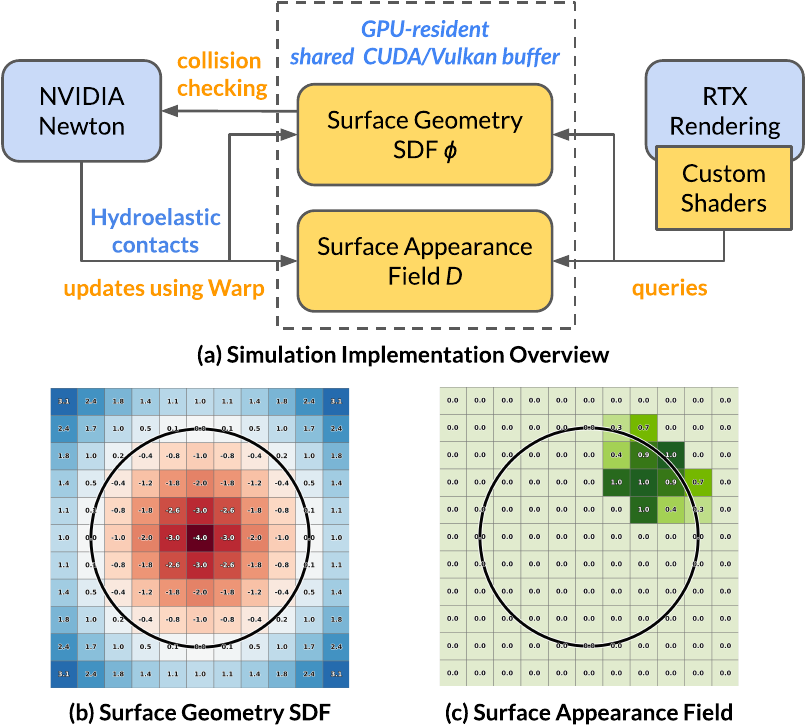}
    \caption{
    Simulation architecture.
    (a) Surface geometry represented by the SDF \(\phi\) and surface appearance
    represented by the field \(D\) reside in a shared CUDA/Vulkan buffer.
    Newton supplies hydroelastic contacts for updates using Warp,
    while custom shaders in Isaac Sim's RTX renderer query the fields directly.
    (b) SDF representation of the workpiece geometry.
    (c) Surface field representing the workpiece surface appearance.
    }
    \label{fig:sim_architecture}
\end{figure}

RoboFin3D is based on Isaac Sim 6.1.0 and the Newton physics engine.
We generate diverse workpieces with sampled welds using WeldGen as inputs
to the simulation. In the simulator, we represent evolving workpiece
geometry by a signed distance field (SDF) \(\phi\) and surface appearance
by a separate surface field \(D\).
Fig.~\ref{fig:sim_architecture}(a) shows the simulation architecture.

\subsection{Welded Workpiece Generation}
\label{sec:weldgen}

WeldGen generates weld beads on the workpieces for simulation and training use.
The input dataset has 8,918 workpiece meshes selected from the ABC dataset~\cite{koch2019abc}.
By repeatedly sampling weld curves at arbitrary locations on a mesh and adding bead geometry,
it can generate an unlimited number of welded workpiece instances.
We vary task difficulty through curve length, geodesic curvature, and surface exposure.
Each instance includes the original workpiece and ground-truth weld geometry
for evaluation and generation of synthetic perception training data.

\subsection{Surface Geometry Evolution for Grinding}
\label{sec:physics_grinding}

The workpiece mesh initializes \(\phi\) in the workpiece frame, with
\(\phi<0\) inside. Its zero level set is the evolving surface
(Fig.~\ref{fig:sim_architecture}(b)).
We compute hydroelastic contact on this field and pass the contacts to
Newton. The hydroelastic model gives a pressure distribution over the
contact patch, which drives local removal.
Removal follows Preston's law~\cite{suratwala2010preston} with a
threshold pressure: the removal rate is proportional to excess pressure and
sliding speed, and no material is removed below the threshold.
The wheel's rotation dominates the sliding speed and runs at a fixed
speed, so we fold the sliding speed into an effective removal
coefficient \(k\):
\begin{equation}
  \partial_t\phi = r\,\lVert\nabla\phi\rVert, \quad
  r(\mathbf{x},t)=k\,\max\bigl(p(\mathbf{x},t)-p_c,\,0\bigr),
  \label{eq:grinding}
\end{equation}
where \(\mathbf{x}\) is position, \(t\) is time, \(r\) is the recession
speed along the surface normal, \(p\) is contact pressure, \(p_c\) is
the threshold pressure, and \(k\) has units of
\(\mathrm{m\,Pa^{-1}\,s^{-1}}\).

\subsection{Surface Appearance Evolution for Sanding}
\label{sec:sanding_appearance}

Sanding removes only a thin coating, so \(\phi\) stays fixed and the
surface field \(D\) records sanding progress
(Fig.~\ref{fig:sim_architecture}(c)). Following Preston's law, \(D\)
accumulates pressure times sliding speed:
\begin{equation}
  D(\mathbf{x},t)=\frac{K_P}{h_c}\int_0^t p(\mathbf{x},\tau)\,
  v_{\mathrm{slip}}(\mathbf{x},\tau)\,d\tau,
  \label{eq:sanding}
\end{equation}
where \(v_{\mathrm{slip}}\) is the sliding speed of the spinning pad,
\(K_P\) is the Preston coefficient of the coating in \(\mathrm{Pa}^{-1}\),
and \(h_c\) is the coating thickness, so the coating is fully removed at
\(D=1\). The renderer blends the coated and exposed appearances, which
differ in color, roughness, and metallic response, with weight
\(\min(D,1)\).

\begin{figure*}[!t]
  \centering
  \includegraphics[width=\textwidth]{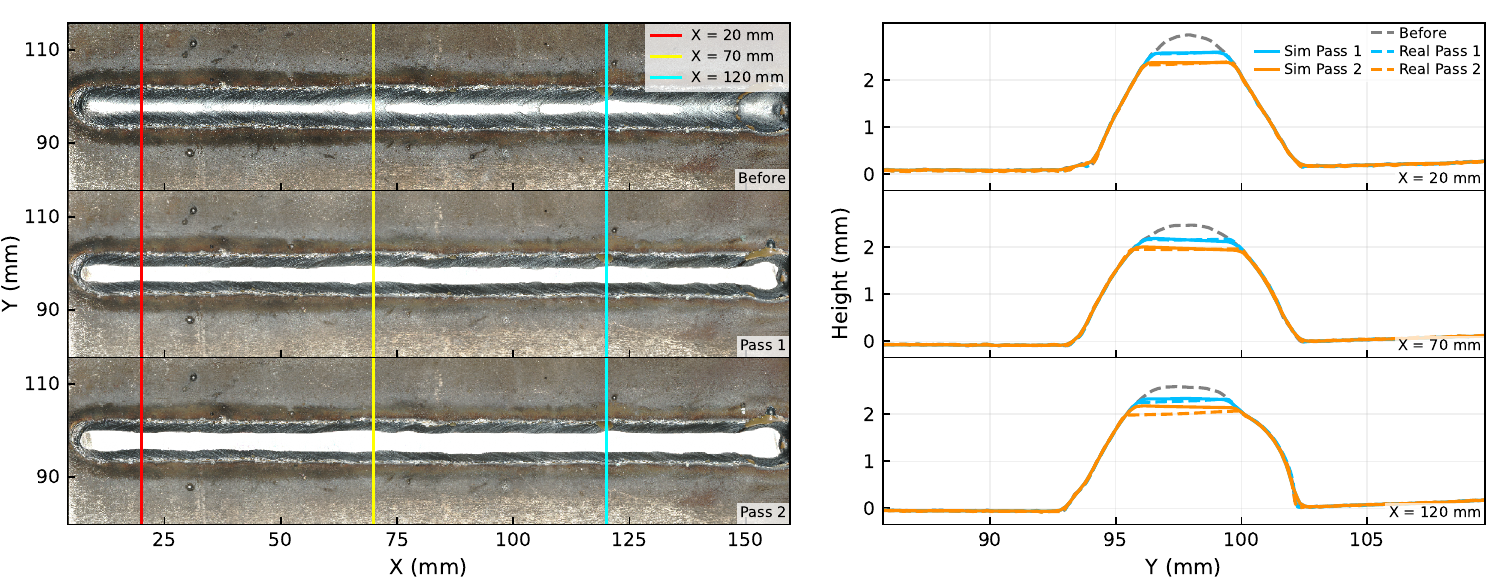}
  \caption{
    Measured and simulated height profiles of a workpiece after two grinding passes.
    Left: optical scans before processing and after each pass,
    with the sections used for comparison marked at $X=20$, $70$, and $120$~mm.
    Right: measured (Real) and simulated (Sim) height profiles at those sections.
    The simulated height profiles resemble the real experimental results.
  }
  \label{fig:grinding_profiles}
\end{figure*}

\subsection{GPU Implementation}
\label{sec:gpu_impl}
The workpiece state is maintained in shared GPU buffers
accessible to CUDA and Vulkan.
Warp updates the fields,
Newton queries the SDF for contact computation,
and custom shaders in Isaac Sim's RTX renderer query the geometry and appearance fields directly.
This fully SDF-native design eliminates
intermediate mesh extraction and remeshing during workpiece evolution,
and avoids CPU readback and unnecessary copies.

\section{Calibration and Perception Transfer}
\label{sec:sim_to_real}

We perform real-to-sim calibration to fit the material removal laws in
Secs.~\ref{sec:physics_grinding} and~\ref{sec:sanding_appearance} using
real grinding and sanding results. For sim-to-real perception, we
fine-tune a segmentation model on automatically labeled simulated
images, either alone or together with a smaller set of labeled real
images, and evaluate it on real sanding test images.

\subsection{Real-to-Sim Calibration}
\label{sec:real_to_sim_calibration}

For both grinding and sanding, we perform two passes of finishing processes
on a workpiece used for calibration in the real world.
We use the first pass to tune contact and removal parameters,
and leave the second pass for evaluation.
The simulation repeats the same processes with the same inputs.
The simulated workpiece is built from an optical scan taken before the first pass,
the tool follows the path of the real pass,
and an admittance controller tracks the same normal force setpoint.
We tune the contact and removal parameters by comparing the simulated surface
with the scan taken after it.

For grinding, we first fit the effective removal coefficient \(k\) to the
volume removed in the first pass. Keeping \(k\) fixed, we then select the
contact stiffness (pressure per unit penetration) and the threshold
pressure \(p_c\) by comparing simulated and measured height profiles.

For sanding, we fit \(K_P/h_c\) by matching
the predicted coating-depletion pattern to the optically estimated exposed substrate.

\subsection{Sim-to-Real Perception}
\label{sec:sim_to_real_perception}

For sanding, we train a perception model on simulated images to segment
the unsanded regions of real workpieces in RGB images. The simulator
provides the labels, so no manual annotation is needed. We render
workpieces at different sanding states, including partially sanded
surfaces and fully sanded ones with no unsanded region left. Within the
rendered workpiece mask, a pixel is labeled sanded if its simulated
appearance has changed from an unsanded reference, and unsanded
otherwise.

The model is SAM2~\cite{ravi2025sam2}, a general segmentation model that
we fine-tune on these images and masks to separate sanded from unsanded
surface. We freeze the pretrained image encoder and train the prompt
encoder and mask decoder. At inference, a workpiece segmentation step
provides the visible workpiece mask and its bounding box. The box
prompts the fine-tuned SAM2 to predict the unsanded region, which is
clipped to the workpiece mask and set to empty when the model predicts
that no unsanded region is present.

\section{Experiments}
\label{sec:experiments}

\subsection{Simulation Evaluation}
\label{sec:simulation_evaluation}

We compare the calibrated simulation with the real passes of
Sec.~\ref{sec:real_to_sim_calibration}. Both processes use a 15~N normal
force setpoint, and the workpieces are scanned before and after each
pass with a Keyence VR-6000 optical profilometer. The simulated second
pass starts from the simulated result of the first pass.

For grinding, both the measured and simulated profiles show
that the rounded crest turns into a plateau in the first pass
and that the second pass lowers it further (Fig.~\ref{fig:grinding_profiles}).
The plateaus differ in height and slope, most after the second pass.
Table~\ref{tab:grinding_removal} compares the cumulative removed volumes.

For sanding, we compare the region where the simulated coating is fully
removed (\(D\geq1\)) with the exposed substrate estimated from the optical
scans. The simulated and measured exposed regions have an intersection
over union (IoU) of 0.81 after the first pass and 0.89 after the second,
and the simulated exposed area is 16.5\% and 11.2\% larger than measured.

\subsection{Perception Evaluation}
\label{sec:perception_evaluation}

The simulated dataset contains 768 images of two workpieces, each in 12 placements.
For each placement, six sanding patterns are captured at five intermediate stages,
plus the unsanded and fully sanded states. The $640\times480$
images are split by placement into 512 training, 128 validation, and
128 test images, so all sanding states of one placement fall in the
same split.

We compare three settings:
\begin{itemize}
\item \textbf{SAM2 Baseline:}
pretrained SAM2 without fine-tuning.
\item \textbf{Sim-only:}
SAM2 fine-tuned on simulated images only.
\item \textbf{Sim + Real:}
SAM2 fine-tuned on simulated images and labeled real images,
which number about 20\% of the simulated training set.
\end{itemize}

All settings use SAM2.1 Tiny with the same post-processing and are
prompted with ground-truth workpiece bounding boxes, so the comparison
excludes workpiece localization errors. Fine-tuning uses 512 AdamW
updates with batch size 4 and learning rate $2\times10^{-5}$, and we
evaluate the final checkpoint without selecting it on the validation
set. We report IoU, Dice, precision, and recall of the unsanded mask
on the real test set.

\begin{figure}[!t]
    \centering
    \resizebox{\columnwidth}{!}{\input{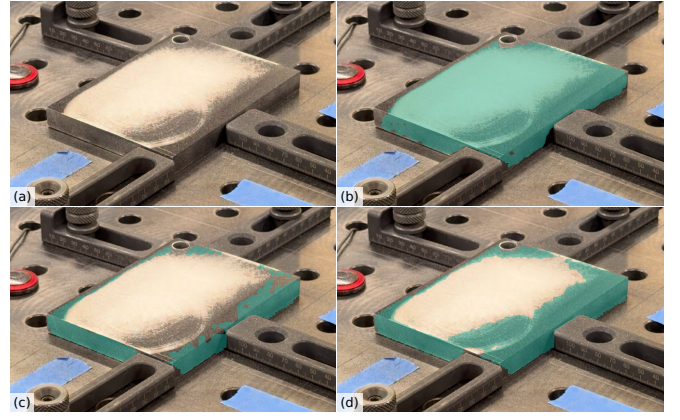}}
    \caption{
  Qualitative results for segmentation of unsanded regions
  on a real workpiece.
  (a) Input RGB image;
  (b) pretrained SAM2;
  (c) Sim-only;
  (d) Sim + real.
  Teal overlays indicate predicted unsanded regions.
  }
    \label{fig:perception_comparison}
\end{figure}

Fine-tuning on simulated images raises IoU from 77.15\% to 84.47\% and
Dice from 86.97\% to 91.42\% (Table~\ref{tab:real_perception}).
Precision rises from 78.94\% to 91.18\% while recall falls from
97.34\% to 91.73\%, so the Sim-only model labels fewer sanded pixels as
unsanded but misses more unsanded pixels. Adding labeled real images
improves all four metrics and raises IoU to 97.41\%.
Fig.~\ref{fig:perception_comparison} compares the three settings on a
real sanding image.

\begin{table}[!t]
    \centering
    \caption{
    \textbf{Cumulative grinding removal.}
    Absolute relative errors use measured volumes as the reference.
    }
    \label{tab:grinding_removal}
    \small
    \setlength{\tabcolsep}{4pt}
    \begin{tabular}{lrrr}
        \toprule
        Pass & Real (mm$^3$) & Sim (mm$^3$) & Error (\%) \\
        \midrule
        1 (calibration) & 139.1 & 111.0 & 20.2 \\
        2 (evaluation)  & 294.8 & 216.1 & 26.7 \\
        \bottomrule
    \end{tabular}
\end{table}

\begin{table}[!t]
    \centering
    \caption{
    \textbf{Segmentation of unsanded regions on real test images.}
    All values are percentages; higher is better.
    }
    \label{tab:real_perception}
    \small
    \setlength{\tabcolsep}{3pt}
    \renewcommand{\arraystretch}{1.1}
    \begin{tabular}{lcccc}
        \toprule
        Fine-tuning data & IoU & Dice & Precision & Recall \\
        \midrule
        None (pretrained)
            & 77.15 & 86.97 & 78.94 & 97.34 \\
        Sim only
            & 84.47 & 91.42 & 91.18 & 91.73 \\
        Sim + Real
            & \textbf{97.41}
            & \textbf{98.68}
            & \textbf{98.59}
            & \textbf{98.79} \\
        \bottomrule
    \end{tabular}
\end{table}

\section{Conclusion}
\label{sec:conclusion}

We presented RoboFin3D, a sim-to-real platform for robotic grinding and sanding.
We use an SDF and a surface field to represent the changing workpiece surface.
WeldGen generates welded workpieces for simulation and perception training.
After calibration, simulated processing effects resemble real ones.
Fine-tuning perception models on simulation-generated data improves their segmentation performance on real workpieces.

Future work includes (i) improving the contact and removal models so
that simulated grinding and sanding match real results more closely,
(ii) speeding up contact computation so that the simulation runs in real
time, (iii) quantifying the sim-to-real gap across more workpieces and
process parameters, and (iv) using RoboFin3D to evaluate grinding and
sanding control policies before real deployment.

\printbibliography[heading=bibintoc]

\end{document}